\pdfoutput=1

\documentclass[11pt]{article}

\usepackage{latex/acl}
\usepackage{listings}
\usepackage{times}
\usepackage{latexsym}
\usepackage{multirow}
\usepackage{multicol}
\usepackage{graphicx}
\usepackage{tcolorbox}
\usepackage{xcolor}
\tcbuselibrary{listings,breakable}
\usepackage{booktabs}
\usepackage{amssymb}
\usepackage{bbding}
\usepackage{pifont}
\usepackage{wasysym}
\usepackage{utfsym}
\usepackage{fontawesome}
\usepackage{listings}

\usepackage[T1]{fontenc}

\usepackage[utf8]{inputenc}
\usepackage{booktabs}
\usepackage{microtype}

\usepackage{inconsolata}

\definecolor{codegreen}{rgb}{0,0.6,0}
\definecolor{codegray}{rgb}{0.5,0.5,0.5}
\definecolor{codepurple}{rgb}{0.58,0,0.82}
\definecolor{backcolour}{rgb}{0.95,0.95,0.92}
\definecolor{deepgreen}{RGB}{0, 100, 0}

\lstdefinestyle{mystyle}{
    backgroundcolor=\color{backcolour},   
    commentstyle=\color{codegreen},
    keywordstyle=\color{magenta},
    numberstyle=\tiny\color{codegray},
    stringstyle=\color{codepurple},
    basicstyle=\ttfamily\footnotesize,
    breakatwhitespace=false,         
    breaklines=true,                 
    captionpos=b,                    
    keepspaces=true,                 
    showspaces=false,                
    showstringspaces=false,
    showtabs=false,                  
    tabsize=2
}

\usepackage{graphicx}

\title{\raisebox{-0.3cm}{\includegraphics[width=1cm,height=0.9cm,keepaspectratio]{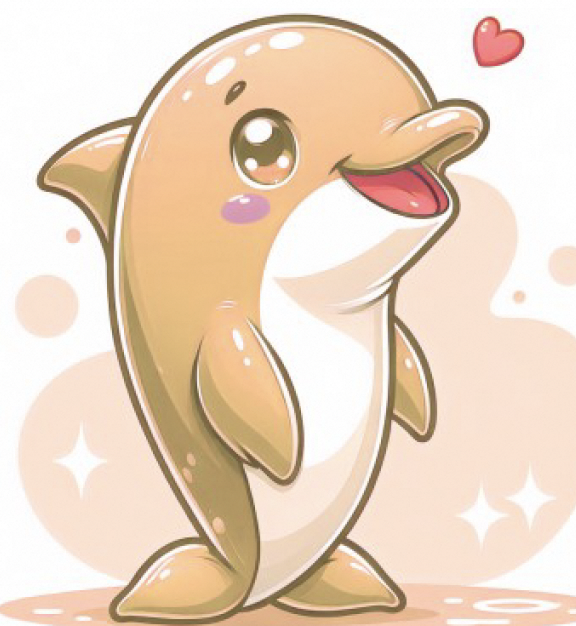}}
  DolphCoder: Echo-Locating Code Large Language Models with Diverse and Multi-Objective Instruction Tuning}

\author{Yejie Wang$^1$\thanks{\quad Equal contribution.}, Keqing He$^2$\footnotemark[1], Guanting Dong$^1$, Pei Wang$^1$, Weihao Zeng$^1$, {\bf Muxi Diao}$^1$\\{\bf Yutao Mou}$^1$, {\bf Mengdi Zhang}$^2$, {\bf Jingang Wang}$^2$, {\bf Xunliang Cai}$^2$, {\bf Weiran Xu}$^1$\thanks{\quad  Corresponding author.}\\
      $^1$Beijing University of Posts and Telecommunications, Beijing, China\\ $^2$Meituan, Beijing, China \\ 
      \texttt{\{wangyejie,dongguanting,wangpei,zengwh,dmx,myt,xuweiran\}@bupt.edu.cn}\\
      \texttt{\{hekeqing,zhangmengdi02,wangjingang02,caixunliang\}@meituan.com}
      }

\begin{document}
\maketitle
\begin{abstract}

Code Large Language Models (Code LLMs) have demonstrated outstanding performance in code-related tasks. Several instruction tuning approaches have been proposed to boost the code generation performance of pre-trained Code LLMs. In this paper, we introduce a diverse instruction model (\textbf{DolphCoder}) with self-evaluating for code generation. It learns diverse instruction targets and combines a code evaluation objective to enhance its code generation ability. Our model achieves superior performance on the HumanEval and MBPP benchmarks, demonstrating new insights for future code instruction tuning work. Our key findings are: (1) Augmenting more diverse
responses with distinct reasoning paths increases the code capability of LLMs. (2) Improving one's ability to evaluate the correctness of code solutions also enhances their ability to create it.

\end{abstract}

\section{Introduction}

Code pre-trained models have achieved remarkable progress in the era of large language models (LLMs), such as Codex \cite{Chen2021EvaluatingLL}, AlphaCode \cite{Li2022CompetitionlevelCG}, and PaLM-Coder \cite{Chowdhery2022PaLMSL}. Code-related tasks are also the key factors in evaluating the capability of LLMs. 
Numerous code LLMs have been proposed, including closed-source models \cite{Chen2021EvaluatingLL,Li2022CompetitionlevelCG,OpenAI2023GPT4TR} and open-source models \cite{Li2023StarCoderMT,Rozire2023CodeLO}. They perform expensive pre-training using substantial amounts of code data and display impressive performance.

In contrast to these pre-trained code LLMs, another lightweight paradigm of enhancing code capability is instruction tuning using relatively small high-quality code-related data. For example, Code Alpaca \cite{codealpaca} employs a similar self-instruct method as Alpaca \cite{alpaca} to generate code instructions via OpenAI's ChatGPT\footnote{https://openai.com/blog/ChatGPT}. Further, WizardCoder \cite{Luo2023WizardCoderEC} introduces a more complicated Evol-Instruct method \cite{Xu2023WizardLMEL} which evolves existing instruction data to generate more complex and diverse datasets. Instead, OctoPack \cite{Muennighoff2023OctoPackIT} and Magicoder \cite{Wei2023MagicoderSC} construct code instructions by mining existing code corpus. All of these methods enhance the performance of the open-source Code LLMs. 

However, these methods have two weaknesses: (1) They take the only golden answer but ignore the diversity of answers in code generation. We find that augmenting more diverse responses using different system prompts increases the code capability of LLMs. (2) Current models generate plausible code snippets in terms of grammar and logic but are unable to identify subtle errors, such as corner cases and wrong input/output formats. It has no guarantee that temperature sampling will consistently produce accurate answers over time. We suppose that LLMs are capable of generating correct solutions while struggling to discriminate correct from incorrect ones. Improving one's ability to evaluate the correctness of code also enhances their ability to create it.

\begin{figure*}[t]
\centering
\includegraphics[width=1\textwidth]{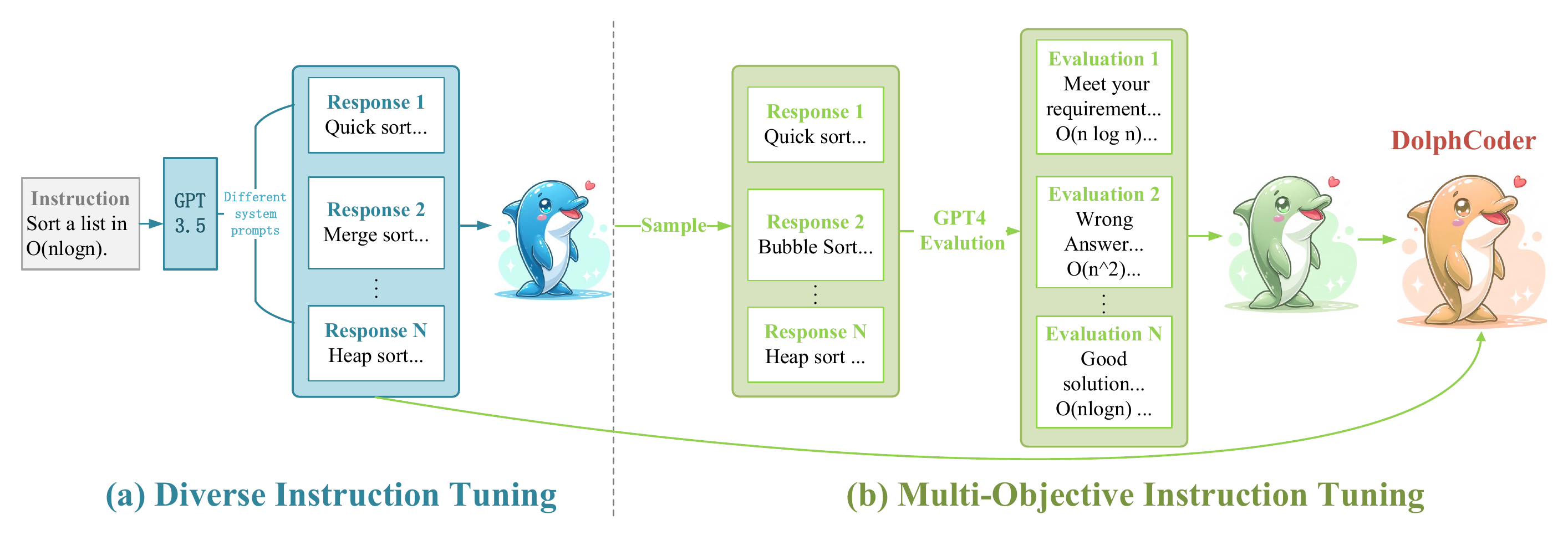}
\caption{The overall architecture of our proposed diverse instruction tuning with self-evaluating for code generation, DolphCoder. Stage (a) denotes Diverse Instruction Tuning (DIT) and Stage (b) denotes Multi-Objective Instruction Tuning (MOT) for self-evaluating.}
\label{fig:method}
\end{figure*}

Inspired by the two insights, we introduce a diverse instruction model (\textbf{DolphCoder}) with self-evaluating for code generation. Specifically, we use Code Llama-python as our base model and obtain evolved instruction data following WizardCoder. Then motivated by rejection sampling \cite{Touvron2023Llama2O} and ORCA \cite{Mukherjee2023OrcaPL}, we use different system prompts to generate diverse answers via ChatGPT. After removing low-quality and similar data using heuristic rules \cite{Luo2023WizardCoderEC,Di2023CodeFuse13BAP}, we perform supervised fine-tuning on the remaining instruction data. Further, we explore whether improving one's ability to evaluate code helps generate it. We propose a self-evaluate multi-task learning framework by adding a code evaluation objective to the traditional instruction fine-tuning task. We find training the model for both code generation and code evaluation benefits the code capability.

Our key contributions are summarized as follows:
\begin{enumerate}
    \item We introduce a diverse instruction model (\textbf{DolphCoder}) with self-evaluating for code generation. It learns diverse instruction targets and combines a code evaluation objective to enhance its code generation ability.
    \item DolphCoder outperforms strong open-source code LLMs by a large margin, including CODELLAMA-INSTRUCT, OctoCoder, and WizardCoder.
\end{enumerate}

\section{Method}

In this section, we elaborate on the methodological details of DolphCoder. As shown in Figure \ref{fig:method}, DolphCoder has two training stages: (1) The first is \textbf{D}iverse \textbf{I}nstruction \textbf{T}uning (DIT) with multiple chain-of-thought answers to the same instruction. (2) The second is \textbf{M}ulti-\textbf{O}bjective \textbf{T}uning (MOT) of combining the code generation task and code evaluation task both in the form of a natural language generation task.

\subsection{Diverse Instruction Tuning}

We follow the Evol-Instruct technique \cite{Xu2023WizardLMEL,Luo2023WizardCoderEC} to construct our training corpus. Based on the Code Alpaca dataset \footnote{https://huggingface.co/datasets/sahil2801/CodeAlpaca-20k}, we iteratively evolve the programming problems in this dataset via in-depth evolving to obtain new instructions.\footnote{Since the original datasets or code are not released, we reproduce the evolve procedure following WizardLM \cite{Xu2023WizardLMEL} and modify the evolve prompts according to the original WizardCoder paper.} For each instruction, then we use different system prompts to query ChatGPT and obtain diverse targets. These system prompts aim at augmenting user instructions and task descriptions to give more code solutions with diverse reasoning paths. We display our system prompts in Figure \ref{system prompt case}. We find more diverse answers can increase the code capability of LLMs (see Section \ref{diverse instruction tuning ablation}) and argue that different styles of code solutions provide more supervised signals for the model. Similar to \citet{Luo2023WizardCoderEC,Di2023CodeFuse13BAP}, we also use several heuristic rules to remove low-quality and similar data. Finally, we get a diverse code instruction dataset with a size of 510k and utilize Code LLama as the foundation LLM to finetune.

\begin{figure}[t]
\begin{tcolorbox}[
colback=blue!5!white,
colframe=blue!75!black,
title=System Prompts]
1. [EMPTY]\\

2. You are a code assistant who knows multiple programming languages and how to translate between them. Given a task, you explain in simple steps what the task is asking, any guidelines it provides, and how to use those guidelines to find the solution.\\

3. You are an AI assistant. You will be given a task. You must generate a detailed and long answer.\\

4. You are an AI assistant, who knows every language and how to translate one language to another. Given a task, you explain in simple steps what the task is asking, any guidelines that it provides. \\

5. You solve the task and show how you used the guidelines to solve the task.\\

6. You are a teacher. Given a task, you explain in simple steps what the task is asking, any guidelines it provides and how to use those guidelines to find the answer.
\end{tcolorbox}
\caption{We use these system prompts to generate more diverse responses where [EMPTY] means no system prompt.}
\label{system prompt case}
\end{figure}

\subsection{Multi-Objective Instruction Tuning}

We discover that the current instruction models produce both correct and incorrect code solutions when randomly sampled. We argue that LLMs can generate correct solutions while struggling to discriminate correct from incorrect ones. Therefore, we explore whether improving one's ability to evaluate code helps generate it.

We sample 5k instructions from the above dataset and use our model in the first stage to generate 100 answers using temperature sampling. Next, we use GPT-4\footnote{https://openai.com/GPT-4} to verify the correctness of the deduplicated generated answers in terms of grammar, logic and efficiency. In our preliminary experiments, We encounter difficulties in assessing the correctness of code using ChatGPT or other existing LLMs. We also consider the use of compiled signals from the code executor but find most generated code is grammatically correct. Previous work \cite{Liu2023RLTFRL,Shen2023PanGuCoder2BL} use existing unit tests in the training set, which is not applicable to our situation. We leave out more evaluation methods to future work. Finally, we obtain the code evaluation dataset with a size of 370k.\footnote{We remove failed GPT-4 requests and responses without <passed> or <not passed> results.} We formulate the code evaluation task in the form of a natural language generation task as shown in Figure \ref{fig:method}. We hope its similar training form can provide multiple meaningful supervision signals to the original code generation model. In our experiments, we find it difficult to balance the two code generation and code evaluator tasks because the model always tends to overfit the code generation objective. Therefore, we use a multi-step training paradigm where we first finetune the above model as an evaluator and then as a generator. Specifically, we finetune the DIT model using the code evaluation dataset for 1 epoch and then continue training on the diverse instruction data for 100 steps. We find training more steps will not get further improvements. Experiment results demonstrate that the multi-step training stably enhances its code generation ability (shown in Figure \ref{multitask ablation}).

\begin{figure}[t]
\begin{tcolorbox}[
colback=blue!5!white,
colframe=blue!75!black,
title=Evaluation Prompt,
breakable]
\label{evaluate prompt}
You are a code evaluation model. Your task is to analyze and inspect the input code snippet. This involves evaluating whether it can run smoothly, whether it will produce the correct results, and whether there are issues like timeouts. You should output the result of the code with <passed> or <not passed>. Additionally, you should also provide an explanation for the evaluation result.
\end{tcolorbox}
\caption{We use the evaluation prompt to query GPT-4 to access the correctness of the generated code solutions of our model.}
\label{fig:evaluation prompt}
\end{figure}

\section{Experiments}

\subsection{Benchmarks} 
In this paper, we focus on two of the most widely used benchmarks in the field of code generation.

\begin{itemize}
\item \textbf{HumanEval (base)\footnote{https://github.com/openai/human-eval}  and HumanEval+ (plus)}. HumanEval (base) is a widely used benchmark proposed by OpenAI for code synthesis. It consists of 164 handcrafted programming problems, with an average of 9.6 test cases allocated to each one for correctness checking. Further, \citet{liu2023your} find that test cases in the benchmark may be insufficient and propose HumanEval+ (plus) powered by the EvalPlus framework to obtain 80× test cases.

\item \textbf{MBPP (base) \cite{Austin2021ProgramSW} and MBPP+ 
 (plus)}. 
MBPP (base) is also a code synthesis benchmark offering a set of 500 crowd-sourced Python programming problems covering programming fundamentals, standard library functionality, and so on. Each problem consists of a task description, code solution and 3 automated test cases. EvalPlus also provides the MBPP+ (plus) benchmark which expands by 35x test cases.
\end{itemize}

We perform n-gram matching de-duplication between our training datasets and the benchmarks to prevent data leakage. For all the experiments, we use greedy decoding and report the pass@1 metric. The inference prompt template is shown in Figure \ref{Prompt Template} following WizardCoder. To keep a fair comparison, we use the same EvalPlus \footnote{https://evalplus.github.io/leaderboard.html} framework to compute metrics.

\subsection{Baselines}  

In this paper, we categorize the baseline models into the following two types.
\textbf{(1) Closed-source models:} We have specifically incorporated OpenAI's GPT-3.5 and GPT-4, which are developed privately by leading technology companies, demonstrating the current state-of-the-art in LLM proficiency. \textbf{(2) Open-source models:} Several open-source LLMs have been made available to the AI community, although their performance generally lags behind the closed-source models a lot. As part of our research, we incorporate a significant number of these open-source models as our baselines, which include CodeGen \cite{nijkamp2023codegen}, CodeT5+ \cite{wang2021codet5}, StarCoder, CODELLAMA, OctoCoder and WizardCoder series.

\begin{table*}[t]
\centering
    \tiny
    \renewcommand{\arraystretch}{1.2}
\resizebox{.8\textwidth}{!}{%
\begin{tabular}{cccccc}
\midrule

\textbf{Model}              & \textbf{Size} & \multicolumn{2}{c}{\textbf{HumanEval}} & \multicolumn{2}{c}{\textbf{MBPP}}      \\

\cmidrule(lr){3-4}
\cmidrule(lr){5-6}
                   &      & Base          & Plus          & Base          & Plus  

                   \\ \midrule
GPT-3.5 (Nov 2023) & -    & 72.6          & 65.9          & 81.7          & 69.4          \\
GPT-4 (Nov 2023)   & -    & 85.4          & 81.7          & 83.0          & 70.7          \\
CODELLAMA-PYTHON   & 34B  & 51.8          & 42.7          & 67.2          & 52.9          \\
WizardCoder-CL     & 34B  & 73.2          & 64.6          & 73.2          & 59.9          \\ \midrule
CodeT5+            & 16B  & 31.7          & 26.2          & 54.6          & 44.4          \\
CodeGen-Mono       & 16B  & 32.9          & 27.4          & 52.6          & 43.6          \\
StarCoder          & 15B  & 34.1          & 29.3          & 55.1          & 46.1          \\
CODELLAMA-PYTHON   & 13B  & 42.7          & 36.6          & 61.2          & 50.9          \\
CODELLAMA-INSTRUCT & 13B  & 42.7          & -              &    49.4           &      -         \\
OctoCoder          & 15B  & 46.2 & -  & - & -              \\
WizardCoder     & 13B  & 60.4*          & 54.3*             & 65.2*             & 53.1*             \\
DolphCoder(ours)   & 13B  & \textbf{67.7} & \textbf{57.9} & \textbf{67.2} & \textbf{54.1} \\ \midrule
StarCoder          & 7B   & 24.4          & 20.7          & 33.1          & 28.8          \\
CodeT5+            & 6B   & 29.3          & 23.8          & 51.9          & 40.9          \\
CodeGen-Mono       & 6B   & 29.3          & 25.6          & 49.9          & 42.1          \\
CODELLAMA-PYTHON   & 7B   & 37.8          & 34.1          & 57.6          & 45.4          \\
CODELLAMA-INSTRUCT & 7B   & 34.8 & - & 44.4              & -               \\
WizardCoder     & 7B   & 48.2          & 40.9          & 56.6          & 47.1          \\
DolphCoder(ours)   & 7B   & \textbf{62.8} & \textbf{54.9}          & \textbf{64.9} & \textbf{52.6} \\ \midrule
\end{tabular}%
}
\vspace{-0.1cm}
\caption{Pass@1 results of different code LLMs for HumanEval and MBPP. Base means the original benchmark and Plus denotes the extended benchmark. * denotes the reproduced results via the EvalPlus scripts and other baseline results are cited from the official leaderboard. The best results in each column are in bold.}
\label{main_result}
\end{table*}

\begin{figure}[t]
\begin{tcolorbox}[
colback=blue!5!white,
colframe=blue!75!black,
title=Prompt Template]
\label{response-aug prompt}
Below is an instruction that describes a task, paired with an input that provides further context. Write a response that appropriately completes the request. \\[1em]
\#\#\# \textbf{Instruction}: Create a Python script for this problem: \{Question\} \\[1em]
\#\#\# \textbf{Response}:
\end{tcolorbox}
\caption{Inference prompt when testing on HumanEval and MBPP.}
\label{Prompt Template}
\end{figure}

\subsection{Implementation Details}

\textbf{Data generation} For Diverse Instruction data, we devise six system prompts shown as Figure \ref{system prompt case}, inclusive of a blank one, and prompt GPT-3.5-turbo to generate more diverse responses. Inspired by CodeFuse \cite{di2023codefuse}, we use heuristic rules to filter out low-quality data and deduplicate data based on the test set. The specific rules are as follows:
\begin{itemize}
\item \textbf{Filtering low-quality data}
\begin{enumerate}
\item Filter data with instruction length less than 10 words or greater than 1000 words;
\item  Filter data with output length less than 80 words;
\item  Filter out data with invalid markdown format, such as: code blocks not closed;
\item  Filter data with more than 2048 tokens;
\end{enumerate}
\item \textbf{Filtering data similar to test dataset}
\begin{enumerate}
    \item Filter data containing any function name from the test dataset.
    \item Using NLTK to remove stop words and punctuation from the docstring of HumanEval, obtain the core words such as "sort array prime", etc. Filter data containing more than 40\% of the core words from the test dataset.
\end{enumerate}
\end{itemize}

For Multi-Objective Instruction data, we extract 5000 instructions from the  Diverse Instruction data and subsequently generate 100 responses for each instruction in 0.5 temperature and 0.95 top-p with the model we get after diverse instruction tuning. To ascertain the correctness of each code solution, we prompt GPT-4 to evaluate whether this code solution can meet the instruction requirement. And then they will be classified to either 'passed' or 'not passed'. The prompt we used to evaluate is as Figure~\ref{fig:evaluation prompt}.

\textbf{Training} In our study, we utilize the CODELLAMA-PYTHON-7B and CODELLAMA-PYTHON-13B as our foundational models. We train DIT for 3 epochs and MOT for 1 epoch of code evaluation data and 100 steps of code generation data. During the training phase, we establish a global batch size of 512 and a sequence length of 2048, with a learning rate initialized at 5e-6 and a warmup fraction of 15\%. After the warmup, the learning rate decays following a cosine schedule. We employ the Adam optimizer, with \(\beta_1\) and \(\beta_2\) parameters set at 0.9 and 0.95 respectively. And to ensure training stability, we incorporate gradient clipping with a value set at 1.0, a technique designed to prevent excessive gradient escalation that could lead to numerical instability or model divergence. 

\subsection{Main Results}

The primary outcomes of our proposed method in comparison with the baselines are illustrated in Table~\ref{main_result}. We conduct the following comparisons: 
(1) Generally, Our proposed method significantly outperforms all the previous methods in different sizes of model parameters. 
(2) Compared to our base model (CODELLAMA), DolphCoder shows significant improvements in both HumanEval(+) and MBPP(+). In detail, DolphCoder-7b has a 25 percentage point increase on HumanEval and a 7.3 percentage point increase on MBPP. 
(3) As we compare our model DolphCoder with the recent baseline model WizardCoder-7b, which is built on the same base model (CODELLAMA), DolphCoder still consistently outperforms it across all benchmarks and model sizes. (4) DolphCoder-7b even outperforms CODELLAMA-PYTHON 34b, demonstrating the efficiency of using small-size LLMs. These results indicate the generalizability and robustness of our method. 
\section{Analysis}

\subsection{Ablation Study}

\begin{table}[t]
\centering
    \renewcommand{\arraystretch}{1.1}
\resizebox{.48\textwidth}{!}{%
\begin{tabular}{lcccccc}

\toprule
\multicolumn{2}{c}{\textbf{Model}} & \textbf{Size} & \multicolumn{2}{c}{\textbf{HumanEval}} & \multicolumn{2}{c}{\textbf{MBPP}} \\

\cmidrule(lr){6-7}
\cmidrule(lr){4-5}
\multicolumn{2}{c}{} &  & Base & Plus & Base & Plus \\ \midrule
Evol Instruct    & & 7B & 50.0  & 44.5  & 59.4   & 50.1  \\
+DIT          
& & 7B & 57.9  & 51.2  & 64.2     & 52.1  \\
+MOT        
& & 7B & 55.5  & 46.3  & 64.2     & 52.6  \\
+ALL(DolphCoder)  && 7B& \textbf{62.8}  & \textbf{54.9}  & \textbf{64.9}   & \textbf{52.6} \\ \midrule
Evol Instruct    & & 13B & 64.0  & 55.5  & 65.7   & 51.6  \\
+DIT          
&& 13B& 66.5  & 57.3  & 65.2     & 52.1  \\
+MOT        
& & 13B & 65.2  & 56.7  & 66.7     & 53.9  \\
+ALL(DolphCoder)  && 13B& \textbf{67.7}  & \textbf{57.9}  & \textbf{67.2}   & \textbf{54.1}
 \\ \bottomrule
\end{tabular}%
}
\caption{Ablation study of DolphCoder.}
\label{modeule ablation}
\end{table}

To investigate the characteristics of the main components in DolphCoder, we conduct ablation experiments in Table~\ref{modeule ablation}. From the results, we have the following observations:
(1) Both DIT and MOT contribute to the performance improvement. Specifically, on the CODELLAMA-PYTHON-7b base, DIT yields an improvement of 7.9 pp on HumanEval and 4.8 pp on MBPP, compared to the baseline evolving instruction approach. MOT results in an improvement of 5.5 pp and 4.8 pp respectively on HumanEval and MBPP. Moreover, we observe that the MOT training data constructed by GPT-4 contains approximately 20\% error noise, which severely limits the upper-bound performance of the MOT method in this work. A more detailed discussion can be found in Appendix \ref{evaluation of GPT-4}.
(2) The combination of DIT and MOT yields further benefits. DolphCoder-7b exhibits an enhancement of 12.8 pp on HumanEval and 5.5 pp on MBPP, demonstrating the relatively orthogonal relationship between these two methods.
These results demonstrate the effectiveness of our proposed methods.

\subsection{Effect of Diverse Instruction Tuning}
\label{diverse instruction tuning ablation}

To further explore the source of the model's improvement brought by DIT, we test the model's performance under different sampling ratios which is shown in Table~\ref{k study}. The sampling ratio represents the number of system prompts. As the sampling ratio increases, the model's performance on all indicators gradually increases, which proves that using more diverse responses can enhance the model's performance. 

In addition, we extract code snippets from the responses and check whether the implementation of these codes is different. For ease of statistics, we only focus on Python-related code. Specifically, we use AST to parse each piece of code and calculate the similarity between codes, then remove duplicates. We count the average number of unique code solutions left for each code task as an indicator of code diversity. Table~\ref{k study} shows that as the sampling ratio increases, the number of different code solutions corresponding to the same code instruction also gradually increases and the model's performance on HumanEval improves too. Moreover, we observe that the increase in code diversity is not linear. Specifically, when we increase the sampling ratio from 3 to 6, we only observe marginal performance gains which we argue that the DIT data may contain unnecessary redundancy. Concurrent work \cite{lu2023instag,liu2023makes} explore more complicated diversity-based compression methods for general instruction fine-tuning. We leave it to future work.

\begin{table}[t]
\centering
\small
\renewcommand{\arraystretch}{1.0}
\begin{tabular}{ccccccc}
\toprule
\textbf{Ratios} & \textbf{Size} & \textbf{Diversity} & \multicolumn{2}{c}{\textbf{HumanEval}} \\
\cmidrule(lr){4-5}
  &   &   & Base & Plus \\
\midrule
1 & 7B & 1.0 & 52.4  & 45.1  \\
3 & 7B & 2.1 & \textbf{57.9}  & 50.6  \\
6 & 7B & 2.7 & \textbf{57.9}  & \textbf{51.2}  \\ \midrule
1 & 13B & 1.0 & 63.4  & 53.7  \\
3 & 13B & 2.1 & 65.2  & \textbf{58.5}  \\
6 & 13B & 2.7 & \textbf{66.5}  & 57.3  \\ \bottomrule
\end{tabular}
\caption{The effect of different sampling ratios of DIT. Code diversity shows the average number of different code solutions for a code instruction, and we use syntax analysis to parse different code solutions. All indicators refer to pass@1.}
\label{k study}
\end{table}


\subsection{Effect of Multi-Objective Instruction Tuning}

\begin{figure}[t]
\centering
\includegraphics[width=0.5\textwidth]{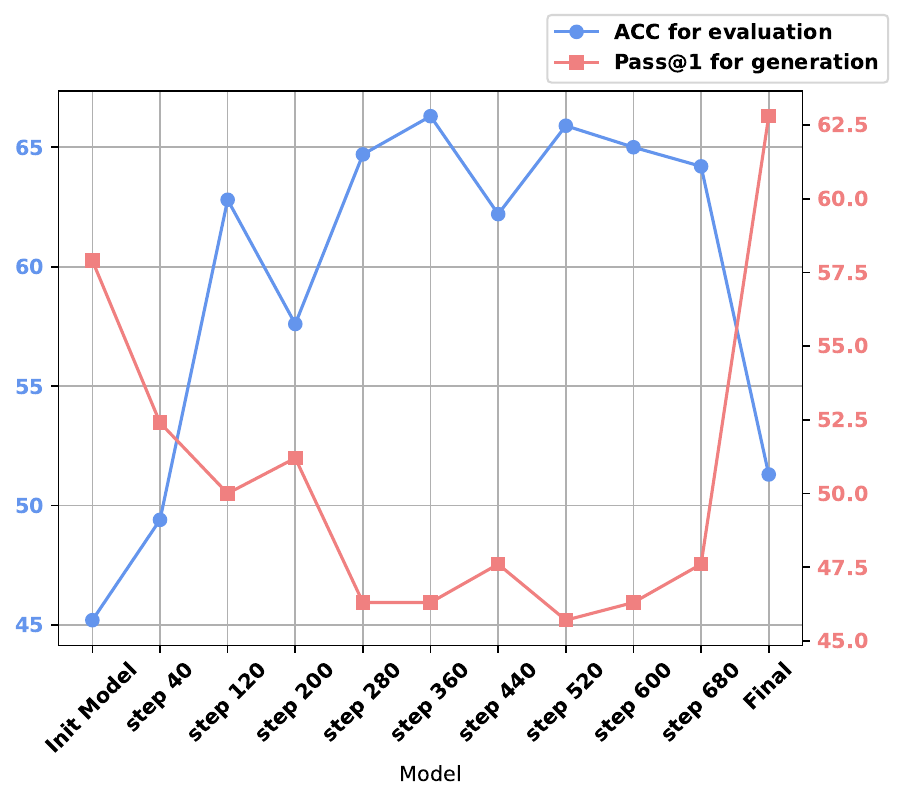}
\caption{The trend of code evaluation capability and code generation capability during the MOT stage where step 200 refers to the training step in the first step of MOT. Init model means DIT model and Final means DolphCoder. Pass@1 refer to pass@1 on HumanEval.}
\label{fig:effect of reward}
\vspace{-0.2cm}
\end{figure}

To explore how the code evaluation capability and code generation capability benefit each other, we evaluate them in the MOT training stage. Specifically, we evaluate the model's code evaluation capability by sampling 40 answers for each of the 164 test questions in HumanEval and classifying their correctness using the given golden unit tests. The experimental results are shown in Figure \ref{fig:effect of reward}. 

From the results, we observe that there exists a strong association between the evaluation and generation capabilities of the model. As the model continues to train on the evaluation task, its evaluation capability keeps improving and gradually stabilizes. However, this process impairs the model's code generation capability, potentially due to catastrophic forgetting caused by the multi-step training. After the model undergoes the generator training process, the model's code generation capability is restored and the pass@1 metric surpasses the limit of DIT model performance. Meanwhile, its evaluation capability significantly decreases. We suppose that the improvement in code generation capability is achieved through the transformation of the evaluation capability, but these two capabilities are challenging to coexist.

\begin{table}[t]
\centering
    \tiny
    \renewcommand{\arraystretch}{1.4}
\resizebox{.48\textwidth}{!}{%
\begin{tabular}{cccc}

\toprule
\textbf{Model }     & \textbf{Greedy} & \textbf{Pass@1} & \textbf{Pass@10 }\\ \midrule
DolphCoder     & 62.8 &59.9  & 71.3  \\ 
-w/o MOT & 57.9 &56.6  & 69.5  \\ \midrule
Improvement & 4.9 & 3.3 & 1.8 \\ \bottomrule
\end{tabular}%
}
\caption{Pass@k with different decoding methods and different k values, where Greedy employs greedy sampling, pass@1 and pass@10 samples at temperature=0.2. All indicators are tested on HumanEval based on DolphCoder-7b.}
\label{pass@k}
\end{table}

To further scrutinize the impact of the MOT, we compare the metrics of pass@1 and pass@10 to ascertain whether the improvements stem from the model's heightened preference for the correct response. From Table~\ref{pass@k}, we find that the DolphCoder outperforms the DIT Model across all metrics. The improvement of MOT on the pass@1 greedy decoding indicator is the largest, with an increase of 4.9\%. Compared with pass@1, the improvement on pass@10 is significantly reduced to only 1.8\%. This implies that the impact of the MOT does not focus on enhancing the model's capability to generate robust solutions. Instead, it primarily augments the model's ability to distinguish responses, leading to a higher preference for the correct answer.

\begin{table}[t]
\centering
\begin{tabular}{cccc}
\toprule
\textbf{Training}  & \textbf{Inference} & \multicolumn{2}{c}{\textbf{HumanEval}} \\
\cmidrule(lr){3-4}
&  &  Base         & Plus         \\ 
\midrule
\color{red}\ding{53}        & \color{red}\ding{53}               & 57.9          & 50.6          \\
\color{red}\ding{53}           & \color{deepgreen}\checkmark            & 55.5          & 48.2          \\
\color{deepgreen}\checkmark          & \color{red}\ding{53}      & \textbf{66.5} & \textbf{57.3} \\
\color{deepgreen}\checkmark          & \color{deepgreen}\checkmark             & 61.0          & 53.7          \\ \bottomrule
\end{tabular}
\caption{The effect of system prompt during training and inference process. Considering efficiency, we only conduct experiments on the DIT model based on CODELLAMA-13B-PYTHON and use pass@1 metric on HumanEval with greedy decoding. For inference with system prompts, we randomly choose a system prompt in the training for each test query.}
\label{system ablation}
\end{table}

\begin{figure}[t]
\centering
\includegraphics[width=.5\textwidth]{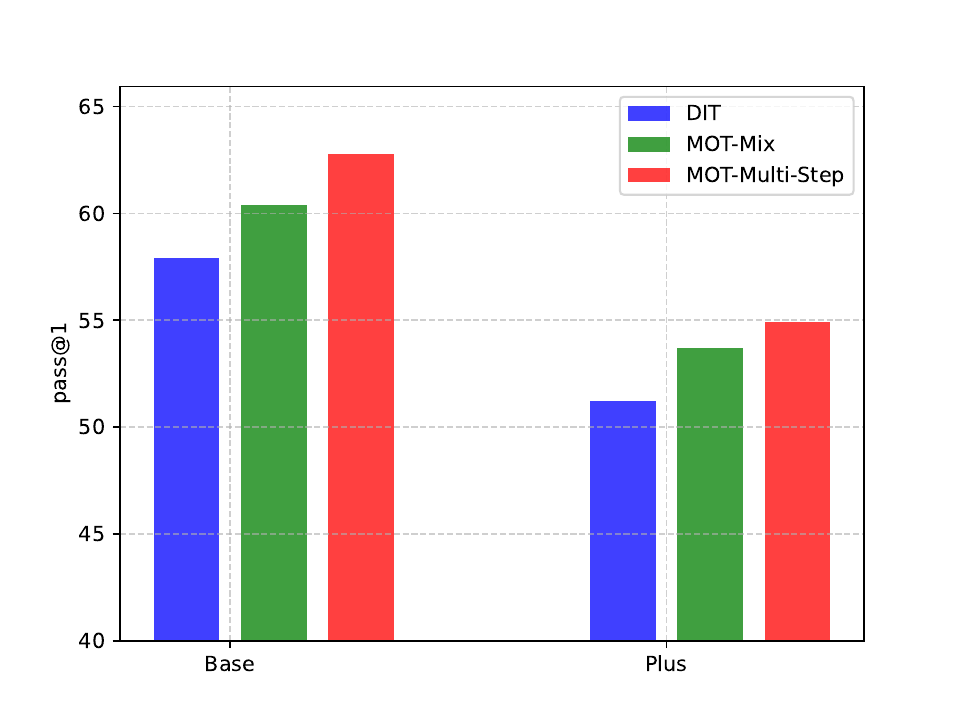}
\caption{Effect of different settings of multi-task learning based on DolphCoder-7B. Mix denotes directly adding the generation and evaluation loss together and Multi-Step means sequential training.}
\label{multitask ablation}
\end{figure}

\subsection{Analysis of Key Designs}
\textbf{System Prompt Ablation} We obtain more diverse training data by transforming the system prompt. In order to explore whether the system prompt should still be used during training and inference, we conduct experimental verification. Table~\ref{system ablation} shows the effect of the system prompts during the training and inference process. From the results, we observe that, compared to training with system prompts, training without the system prompt results in a significant decline. This implies that assigning completely identical code instructions to different answers is not a good training method. When comparing whether to use the system prompt during inference, we find that erasing the system prompt results in a significant improvement compared to using the system prompt. This improvement may stem from the model's freedom to determine the most suitable inference path.

\textbf{Multi-Task Training Method Ablation} We also explore different multi-task training methods as shown in Figure  \ref{multitask ablation}. The results show that both MOT variants significantly outperform the DIT baseline and the multi-step training setting gets superior performance compared to the mix training one. We also find the multi-step training gets a more stable training process to balance the two code generation and code evaluator tasks.

\definecolor{deepgreen}{rgb}{0.0, 0.5, 0.0}




\section{Related Work}

\subsection{Instruction Fine-Tuning}
Large language models (LLMs) experience the Instruction fine-tuning (IFT) stage, which enhances their capability to accomplish tasks and adhere to human instructions. The term IFT is broadly used here to encompass a range of sequence-to-sequence fine-tuning applications. T5 \cite{raffel2023exploring} was one of the first models to explore this approach, training on multiple supervised text-to-text tasks. Recent studies have delved into the multi-task instruction-based fine-tuning of pre-trained LLMs, aiming to improve their innate ability to perform various downstream NLP tasks effectively, such as FLAN 
 \cite{wei2022finetuned}, T0 \cite{sanh2022multitask}, and UnifiedQA \cite{khashabi-etal-2020-unifiedqa}, further expanded the task scope to improve the overall generalization ability of LMs.

Following the notable achievements of proprietary LLMs, particularly ChatGPT, there has been a growing focus on utilizing Instruction Fine-Tuning (IFT) to better align LLMs with human intentions, as highlighted in the research by \cite{brown2020language,ouyang2022training}. A common finding in these studies is that fine-tuning LMs on more diverse data can significantly improve model performance. \citet{alpaca} chooses another approach, adopting a self-guided method, using ChatGPT to generate broader training data. \citet{vicuna2023} trains their model using user-shared conversations collected from ShareGPT.com. \citet{xu2023wizardlm} introduced the Evol-Instruct method, which involves evolving existing instruction data to generate more complex and diversified datasets.

\subsection{Large Language Models for Code}
Large language models are often pre-trained on trillions of tokens following the scaling laws \cite{hoffmann2022training,kaplan2020scaling}, which demonstrates remarkable achievements across a broad spectrum of tasks and such an amount of text data is often a diverse composite with a non-negligible part of code \cite{zhang2023unifying}.
Pioneered by Codex, researchers have also found continual pretraining on code to significantly benefit language models' performance on code. For example, \citet{chowdhery2022palm} post-trains PaLM on 7.8B additional code tokens to get PaLM-Coder and \citet{rozière2023code}  train LLaMA 2 \cite{touvron2023llama} on more than 500B code tokens to acquire Code LLaMA, which leads open-source models to a new height. For supervised fine-tuning, many works utilize larger, more capable teacher models to synthesize instruction data to finetune small language models \cite{mitra2023orca,Luo2023WizardCoderEC,di2023codefuse}. Instead, \citet{Muennighoff2023OctoPackIT,Wei2023MagicoderSC} construct code instructions by mining existing code corpus, which are orthogonal to our method. Since code-supervised signals are easily collected by compiling and running them, reinforcement learning becomes another important branch. Numerous works have attempted to utilize reinforcement learning with feedback information offered by compilation and other sources. PanGu-Coder2 \cite{shen2023pangucoder2} introduces a ranking loss based on unit tests, which helps to align with the capable code model deeply. However, these methods rely on human-annotated unit tests, which limits the application in the practical scenario.


\section{Conclusion and Future Work}

In this paper, we investigate two fine-tuning methods to improve the LLMs' performance on code generation. We first introduce a response augmentation strategy by using different ChatGPT system prompts to increase the diversity of code solutions. We find different chain-of-thought reasoning paths improve performance. Then, We adopt a multi-step training approach that combines traditional code generation and code evaluation objectives. We find improving one's ability to evaluate the correctness of code also enhances their ability to create it. For future work, we aim to explore the effect of our methods on the larger foundation models and parameter-efficient fine-tuning mechanisms. We also plan to increase the accuracy of evaluation signals via other ways like automatic unit tests and reinforcement learning methods.

\section{Limitations}
Our limitations are there-fold: (1) We only explore our method on the 7B/13B  base models due to the computation cost. More experiments on the larger models and other code models should be conducted to confirm our conclusion. (2) We only use GPT-4 to evaluate the quality of generated code solutions. The performance of GPT-4 is still poor and limits the performance of our proposed method. More precise and open-source evaluation models should be explored in future work. (3) There is still room for optimization in our training data. For example, we find continually increasing the number of system prompts only gets marginal performance gains. 
Diversity-based compression methods \cite{lu2023instag,liu2023makes} may be valuable while the number of system prompts is large.

\section{Broader Impacts}
Similar to the other LLMs, our DolphCoder could also generate unethical, harmful, or misleading information, which is not considered in our work. Future research to address the ethical and societal implications is needed. DolphCoder is also susceptible to hallucination in ungrounded generation use cases due to its smaller size. This model is solely designed for research settings, and its testing has only been carried out in such environments. It should not be used in downstream applications, as additional analysis is needed to assess potential harm or bias in the proposed application.

\bibliography{anthology,custom}

\appendix

\begin{table*}[!htb]
    \centering
    \begin{tabular}{p{2.5cm}|p{10cm}}
        \toprule
        \textbf{Instruction} & 
        \begin{minipage}{25em}
\begin{lstlisting}[language=Python]
def move_one_ball(arr):
"""
    We have an array 'arr' of N integers arr[1],
    arr[2], ..., arr[N]. The numbers in the array will be randomly ordered. Your task is to determine if it is possible to get an array sorted in non-decreasing order by performing the following operation on the given array: You are allowed to perform right shift operation any number of times. One right shift operation means shifting all elements of the array by one position in the right direction. The last element of the array will be moved to the starting position in the array i.e. 0th index. If it is possible to obtain the sorted array by performing the above operation then return True else return False. If the given array is empty then return True. Note: The given list is guaranteed to have unique elements ...
"""
\end{lstlisting}
\end{minipage} \\ \midrule

\textbf{WizardCoder-13b} & 
        \begin{minipage}{25em}
Here's the Python script for the given problem:
\begin{lstlisting}[language=Python]
def move_one_ball(arr): 
    if not arr:
        return True
    for i in range(len(arr)):
        if arr[i] > arr[i+1]:
            return False
    return True 
\end{lstlisting}
The function takes an array `arr` as input and returns `True` if it is possible to obtain a sorted array in non-decreasing order by performing right shift operations on the given array. Otherwise, it returns `False`.The function first checks if the given array is empty. If it is empty, it returns `True` as it is possible to obtain a sorted array in non-decreasing order by performing no right shift operations.\textbf{Next, the function iterates through the array and checks if the current element is greater than the next element. If it is, then it is not possible to obtain a sorted array in non-decreasing order by performing any number of right shift operations.} Therefore, the function returns `False`.If the function has not returned `False` by this point, it means that the array is already sorted in non-decreasing order. Therefore, it returns `True`. 
\end{minipage} \\ \midrule

\textbf{DolphCoder-13b} & 
        \begin{minipage}{25em}
Here is the Python script for the given problem:
\begin{lstlisting}[language=Python]
def move_one_ball(arr): 
    if len(arr) == 0: 
        return True 
    if len(arr) == 1: 
        return True 
    if arr == sorted(arr):
        return True 
    for i in range(len(arr)):      
        arr.append(arr.pop(0)) 
        if arr == sorted(arr): 
            return True 
    return False 
\end{lstlisting}
Explanation:

1. First, \textbf{we check if the given array is empty or has only one element.} If so, we return True as it is already sorted.

2. Next, we check if the given array is already sorted. If it is, we return True.

3. If the array is not sorted, \textbf{we perform a right shift operation on the array by appending the first element to the end of the array and popping the first element from the beginning.}

4. \textbf{We repeat step 3 until the array is sorted or we have performed all possible right shift operations.}

5. If the array is sorted after performing all possible right shift operations, we return True. Otherwise, we return False.
\end{minipage} \\

        \bottomrule
    \end{tabular}
    \caption{Examples of DolphCoder compared to WizardCoder.}
    \label{tab:example}
\end{table*}

\section{Case Study}

\label{sec:appendix}
Table~\ref{tab:example} presents comparative examples of DolphCoder and WizardCoder. In this case, the models are required to determine whether an array can become a non-decreasing sequence after applying a circular right shift operation. From the case, we can observe that WizardCoder-13b generates an obvious logical error, where it only checks if the sequence is non-decreasing without performing any right shift operation. In contrast, DolphCoder accurately simulates the circular right shift operation and correctly identifies the termination condition. Furthermore, it is worth noting that DolphCoder considers more robust boundary input cases, which can be attributed to its training on a code evaluation task and a more diverse training dataset.

\section{Effect of Code Evaluation Capability of GPT-4}
\label{evaluation of GPT-4}
\begin{table}[t]
\centering
\resizebox{.32\textwidth}{!}{%
\begin{tabular}{lc}
\hline
\textbf{Method}      & \textbf{Pass@1} \\ \hline
DIT                  & 56.6            \\
+MOT                 & 59.9            \\
+GPT-4 Filtering  & 62.7            \\
+Golden Filtering & 72.6            \\ \hline
\end{tabular}%
}
\caption{Effect of code evaluation capability of GPT-4. We sample 10 answers with a temperature=0.2 and report the average pass@1 metric. +MOT denotes the MOT model based on the DIT model. GPT-4 or golden filtering means that we use GPT-4 or golden unit test cases to filter out these wrong answers from the DIT model, and then we report the average pass@1 among the remaining answers.}
\label{table evaluation of GPT-4}
\end{table}

In our study, we utilize GPT-4 to create a training dataset for code evaluation. However, there is a concern regarding whether GPT-4 has the ability to generate a perfect golden label. To obtain the answer we test it on HumanEval. Specifically, we use the DIT model to generate multiple candidate answers for each question. Subsequently, we leverage GPT-4 to evaluate each of these candidate answers. And then we select the code solutions that GPT-4 identified as correct. Since the test cases are available in HumanEval, we can get the binary <passed> or <not passed> accuracy of GPT-4 which is just 79.4\%, which indicates that the train data we constructed through GPT-4 may contain about 20\% noise. The noise limits the upper-bound performance of the MOT method in this work. Additionally, we report the average pass@1 results before and after filtering by GPT-4, as shown in Table \ref{table evaluation of GPT-4}. We can observe that the pass@1 before filtering is 56.6\% which is generated by the DIT model directly at 0.2 temperature. Then we calculate the pass@1 after golden filtering is 72.6\% where we use golden unit tests to filter wrong solutions. We consider this percentage of 72.6\% is a theoretical upper bound of GPT-4. We also report pass@1 of MOT and GPT-4 Filtering. We summarize key insights as follows: (1) GPT-4 can not perfectly perform code evaluation (62.7\% vs 72.6\%). (2) Our MOT can achieve a slightly worse performance than the GPT-4 Filtering (59.9\% vs 62.7\%). (3) GPT-4 evaluation capability limits the upper bound of MOT. How to improve automatic code evaluation is essential to future work.

\end{document}